\documentclass[conference]{IEEEtran}
\IEEEoverridecommandlockouts
\usepackage{cite}
\usepackage{amsmath,amssymb,amsfonts}
\usepackage{algorithmic}
\usepackage{graphicx}
\usepackage{textcomp}
\usepackage[table]{xcolor}
\usepackage{booktabs}
\usepackage{url}
\usepackage{array}
\usepackage{xspace}

\makeatletter
\newcommand{\linebreakand}{%
  \end{@IEEEauthorhalign}%
  \hfill\mbox{}\par
  \mbox{}\hfill\begin{@IEEEauthorhalign}%
}
\makeatother

\def\BibTeX{{\rm B\kern-.05em{\sc i\kern-.025em b}\kern-.08em
    T\kern-.1667em\lower.7ex\hbox{E}\kern-.125emX}}
\begin{document}

\title{BrailleBench: Investigating Multi-Criteria Braille Comprehension in Large Language Models}

\author{\IEEEauthorblockN{Jinghan Zhang}
\IEEEauthorblockA{\textit{School of Computing} \\
\textit{Clemson University}\\
Clemson, USA \\
jinghaz@clemson.edu}
\and
\IEEEauthorblockN{Fengran Mo}
\IEEEauthorblockA{\textit{Department of Computer Science} \\
\textit{Rochester Institute of Technology}\\
Rochester, USA \\
frmvcs@rit.edu}
\and
\IEEEauthorblockN{Zhiyu Chen}
\IEEEauthorblockA{\textit{Amazon.com, Inc.}\\
Seattle, USA \\
zhiyuche@amazon.com}
\linebreakand
\IEEEauthorblockN{Xiaoyan Han}
\IEEEauthorblockA{\textit{School of Computing} \\
\textit{Clemson University}\\
Clemson, USA \\
xiaoyah@clemson.edu}
\and
\IEEEauthorblockN{Kunpeng Liu}
\IEEEauthorblockA{\textit{School of Computing} \\
\textit{Clemson University}\\
Clemson, USA \\
kunpenl@clemson.edu}
\and
\IEEEauthorblockN{Chang-Tien Lu}
\IEEEauthorblockA{\textit{Department of Computer Science} \\
\textit{Virginia Tech}\\
Alexandria, USA \\
ctlu@vt.edu}
}

\maketitle

\begin{abstract}
Although Large language models (LLMs) mediate access to knowledge and computational assistance, their capabilities should benefit vulnerable groups in the same way.
However, it is unclear whether existing AI systems are inclusive enough for blind and deafblind users to access the same functionality through Braille, whose indicators, contractions, and digital representations introduce distinct requirements for model comprehension.
To this end, we introduce BrailleBench, a benchmark for evaluating LLMs in Braille comprehension from different Criteria. BrailleBench aligns 5,570 instances from five datasets, including mathematics, commonsense, and multi-hop question answering across English and Braille Grades~1 and~2. Different configurations are designed to understand whether the systems can comprehend Braille-authored content, express answers in Braille, and complete end-to-end Braille interaction. 
To ensure the quality and prevent evaluation bias, the benchmark is built through a deterministic, expert-reviewed pipeline via a self-created Braille Toolkit without using any data instances generated by LLMs.
We evaluate six representative LLMs from various aspects. 
The results reveal a persistent gap between print-English capability and Braille accessibility. Braille understanding and expression are asymmetric, where Grade~2 is especially fragile on the input side compared to Grade~1, and fully Braille requests further reduce performance. 
The experimental observations provide valuable guidance for the development of future Braille AI systems. 
All related resources in BrailleBench are publicly available for future research.
\end{abstract}


\section{Introduction}
The development of large language models (LLMs) in the era of generative AI should be inclusive to benefit everyone.
These LLM-based systems mediate access to knowledge and computational assistance~\cite{mo2026opendecoder}, enabling common users to support their daily lives.
However, it is unclear whether users can access the same functionality through disability-related reading and writing systems, such as Braille for blind and deaf-blind users.
This is important for making AI systems truly inclusive.

In terms of the interaction between blind users and AI systems, Braille is not merely an alternative rendering of speech or print. 
It serves as a private, editable input channel for precise technical data (e.g., natural language, code, mathematics, and notation) when speech interaction becomes impractical due to ambiguity or inefficiency.
Supporting Braille in AI systems is therefore a question of interaction accessibility rather than simply adding another text format, which is underexplored in existing LLM-based systems. Earlier AI research on Braille has primarily addressed recognition and translation through rule-based, natural-language-processing, and neural approaches~\cite{ali2023aiBraille,li2017braillesketch,zhang2024prototypical}; whether general-purpose LLMs can sustain task-oriented interaction through Braille remains underexplored.


In this study, we aim to understand the capability of existing LLMs to comprehend Braille-authored messages by evaluating them under credible and practical criteria.
To this end, we first identify three different scenarios with accessibility through a user's reading and writing interface: (i) Braille to English to measure access to Braille-authored content, e.g., accessing reading, (ii) English to Braille to measure accessible response generation, e.g., accessing writing, and (iii) Braille to Braille to measure end-to-end interaction in Braille.
Two complementary English Braille forms\footnote{Different languages may have their own corresponding Braille systems, while our study focuses only on English-related Braille.}, Grade 1~\cite{icebUEB2024} and Grade 2~\cite{rnibContractedBraille} Braille, are included to distinguish basic uncontracted or contracted Braille capability from the higher level requirements of contracted Braille used in fluent reading. Grade~1 represents words without contractions, while Grade~2 replaces frequent words and letter groups with shorter but context-dependent signs, which requires more complex lexical and positional disambiguation.


Intuitively, strong performance through original English-to-English interaction does not guarantee usable performance in Braille settings.
The challenge for LLMs to comprehend Braille might lie in recognizing the Braille pattern in a consistent sequence rather than isolated character or word transcription. 
As shown in Figure~\ref{fig:representation-examples}, the way for blind people to access mathematical knowledge should be in an ASCII sequence rather than a structured equation in print English, which requires the LLMs to comprehend in the same way. Besides, different Braille standards in Grade 1 and Grade 2 will produce various semantic mapping results when given the same original input sequences.
Since these capacities might not be involved in existing LLMs' pre-training procedure, comprehensive Braille interactions cannot be guaranteed~\cite{wang2025diversity,zhang2026starpo}.

To analyze such a capacity of LLMs, we construct \textbf{BrailleBench}, a reproducible and comprehensive evaluation framework 
to evaluate Braille comprehension, task completion, and accessible expression. 
Specifically, Braille comprehension and Braille generation are distinct and asymmetric capabilities.
This is because, on the input side, a model should recover the intended linguistic and symbolic content. On the output side, it should follow the requested Grade representation rather than return print English as default and avoid code-switch~\cite{mohamed2026lost}. 
To prevent the content in the benchmark from being generated by LLMs that would result in evaluation bias risk, we build a toolkit based on \texttt{liblouis} and fixed UEB tables to control the Braille expression quality.
We evaluate six representative LLMs with detailed quantitative analysis. 
Our investigation shows that the existing LLMs perform in different implicit patterns when encountering a Braille environment, which should be related to the reasoning mechanisms~\cite{zhang2025ratt,zhang2025entropy,zhang2026blind} and whether Braille is included in the pre-training corpus. These findings provide valuable guidance for the development of future Braille AI systems.
The whole benchmark\footnote{BrailleBench: \url{https://github.com/jinghanzhang1998/braillebench}} and toolkit\footnote{
Braille Toolkit: \url{https://github.com/jinghanzhang1998/braille-toolkit}} for constructing Braille are available at our public repositories.


\begin{figure}[t]
\centering
\renewcommand{\arraystretch}{1.16}
\begin{tabular}{
    @{}>{\bfseries}p{0.14\columnwidth}
    p{0.78\columnwidth}@{}
}
\toprule
\multicolumn{2}{c}{\textbf{(a) Mathematical expression}} \\
\midrule
Print EN &
\begin{minipage}[t]{\linewidth}
Evaluate the limit as \(x\) approaches \(0\):
\[
\lim_{x\rightarrow 0}
\frac{\sqrt{1+x+x^{2}}-\sqrt{1-x+x^{2}}}{\sin x}.
\]
\end{minipage}
\\
G1 &
{\scriptsize\ttfamily\raggedright
,evaluate the limit as x approaches \#j3
"<sqrt"<\#a "6 x "6 x`5\#b"> -
sqrt"<\#a - x "6 x`5\#b">">\_/
"<sin"<x">">4
\par}
\\
\midrule
\multicolumn{2}{c}{\textbf{(b) Natural-language question}} \\
\midrule
EN & Who is older, Annie Morton or Terry Richardson? \\
G1 & {\ttfamily ,who is older1 ,annie ,morton or ,terry ,richardson8} \\
G2 & {\ttfamily ,:o is old\}1 ,annie ,morton or ,t\}ry ,ri*>dson8} \\
\bottomrule
\end{tabular}

\caption{Two examples to illustrate the challenges in Braille-based interaction.
(a) How print mathematics will be conveyed into Grade~1 Braille as a longer sequence of cells and indicators.
(b) The same question will be expressed in different sequences in terms of Grade~1 and Grade~2 based on contractions.}
\label{fig:representation-examples}
\end{figure}

Our contributions are summarized as follows:
\begin{enumerate}
    \item We construct a benchmark with 5,570 instances for Braille comprehension, which is crucial to facilitate inclusive interaction between blind users and AI systems and is overlooked in the literature. Both Grade~1 and Grade~2 Braille and different types of tasks are included. 
    \item We build a toolkit \textsc{Braille Toolkit} based on \texttt{liblouis} and fixed UEB tables to control the Braille expression quality, which prevents evaluation bias and ensures reproducibility.
    \item We conduct a comprehensive evaluation of six representative LLMs. 
    The analysis covers a range of scenarios to reveal the implicit pattern behind different LLMs for Braille comprehension, which is useful for the development of future Braille AI systems.
\end{enumerate}

\section{Related Work}

\subsection{Challenges of Braille Comprehension in Language Models}
\label{sec:settings}

Digital Braille can be presented to language models as Unicode patterns, dot-number notation, or Braille ASCII. Unicode assigns dedicated code points to Braille cells~\cite{unicodeBraille}, but rare symbols remain poorly matched to English-centric tokenizers~\cite{velayuthan2025tokenizers,ma2023comprehensive}; dot notation expands each cell into several digits and delimiters, while rendered Braille additionally requires a visual encoder. Braille ASCII instead represents each six-dot cell with one printable character and is used in Braille Ready Format files~\cite{locBRF}. It can pass through standard text APIs without rendering and retains compact ASCII sequences, although familiar characters must be interpreted as Braille cells rather than print English. This compatibility neither implies one model token per cell nor makes ASCII a universal user interface; it only reduces rendering and interface barriers while preserving the Braille-decoding problem. We therefore use Braille ASCII as the primary interface and test Unicode and dots as alternative surface forms.

Unified English Braille (UEB) distinguishes uncontracted Grade~1 from contracted Grade~2~\cite{icebUEBOverview}. Grade~1 preserves words letter by letter and is commonly taught first~\cite{icebUEB2024}; Grade~2 replaces frequent words and letter groups with shorter signs and is widely used by experienced readers and in longer texts~\cite{rnibContractedBraille}. For LLMs, Grade~1 is longer but keeps spelling explicit, whereas Grade~2 is shorter but requires lexical and positional disambiguation. Evaluating both grades therefore separates sequence-length difficulty from the contextual difficulty of contractions while covering both spelling-explicit and fluent-reading use.

Prior accessible-interaction systems also position blind users as active authors rather than passive recipients: BrailleSketch enables gesture-based Braille text entry, while ADCanvas combines a screen-reader-accessible editor with a multimodal LLM agent for blind and low-vision creators~\cite{li2017braillesketch,li2026adcanvas}. These systems motivate evaluating both access to user-authored input and accessible model output.

\subsection{LLM-Based Braille Systems}

Existing language-model work on Braille primarily studies translation, disambiguation, and visual recognition. Low-resource Chinese Braille systems address data scarcity and structural ambiguity through Braille-specific pre-training, joint translation--segmentation, dedicated tokenization and generation modules, or curriculum learning over synthetic and tone-omitted data~\cite{yu2023pretraining,huang2024joint,chen2026mt5,wu2024visionbraille,zhang2025leka}. Prompted LLMs have been used to resolve pronunciation choices in Taiwanese Mandarin and Japanese Braille~\cite{watty2023prompt,kitsunai2024japanese}, while context-aware generation with automated back-translation targets text-to-Braille fidelity~\cite{an2026textbraille}. For technical content, Gross et al.~\cite{gross2026mathbraille} evaluate LLMs on bidirectional conversion between MathML and Nemeth or UEB. General-purpose models nevertheless show inconsistent comprehension and generation of Unicode Braille~\cite{daniel2024unicode}, and multimodal models can propagate optical transcription errors into downstream answers~\cite{karamolegkou2025visualassistants}. BrailleLLM~\cite{huang2025braillellm} uses instruction tuning for translation, formula conversion, and Braille question answering; Korean experiments~\cite{abdullah2026braille} similarly find task-specific supervised models more reliable than unadapted LLMs. Together, these studies establish the difficulty of zero-shot Braille processing and the value of adaptation, but largely evaluate translation, conversion, or recognition. Our BrailleBench instead holds the task and target fixed while separately measuring text-only Braille reading, Braille response generation, and end-to-end interaction in Grade~1 and Grade~2, without assuming Braille-specific model adaptation.

\section{BrailleBench}
\label{sec:benchmark}
\begin{figure*}
    \centering
    \includegraphics[width=\linewidth]{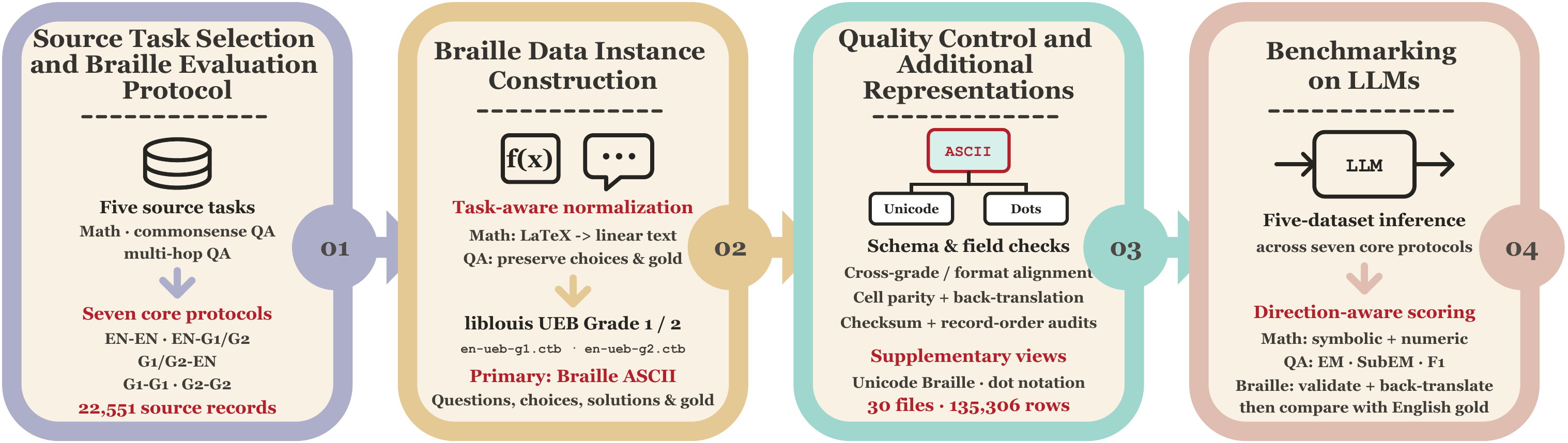}
    \caption{Overview of the BrailleBench construction and evaluation pipeline. Five source datasets are adapted to seven English--Braille interaction protocols, normalized by task type, and deterministically transcribed into UEB Grade~1 and Grade~2 Braille with \texttt{liblouis}; the aligned Braille ASCII records are then validated, supplemented with Unicode and dot-number representations, and evaluated on LLMs using direction-aware scoring.}
    \label{fig:placeholder}
\end{figure*}
To evaluate whether LLMs can comprehend Braille in task-oriented interactions, we construct the \textbf{BrailleBench} benchmark. 
The construction pipeline contains four stages: (i) Source Task Selection and Braille Evaluation Protocol; (ii) Data Sample Construction; (iii) Quality Control Mechanism; and (iv) Benchmarking on LLMs, as shown in Figure~\ref{fig:placeholder}.

\subsection{Source Task Selection and Braille Evaluation Protocol}
\label{sec:source-tasks}
The data instances in BrailleBench are inherited from five widely used datasets, including grade-school arithmetic, competition mathematics, commonsense reasoning, and multi-hop question answering tasks, which are shown in Table~\ref{tab:dataset-stats}.
There are a total of 5,570 source instances, where we leverage the whole test set of GSM8K~\cite{cobbe2021gsm8k}, AIME~2024~\cite{mathai_aime24}, and CommonsenseQA~\cite{talmor2019commonsenseqa}. For HotpotQA~\cite{yang2018hotpotqa} and 2WikiMultiHopQA~\cite{ho2020twoWiki}, the first 1,500 examples in each of the development sets are used to bound the total cost.


\begin{table}[t]
\centering
\begin{tabular}{
    @{}>{\raggedright\arraybackslash}p{0.32\columnwidth}
    >{\raggedright\arraybackslash}p{0.36\columnwidth}
    >{\raggedright\arraybackslash}p{0.23\columnwidth}@{}
}
\toprule
\textbf{Dataset} & \textbf{Task} & \textbf{Answer Format} \\
\midrule
GSM8K & Grade-school arithmetic & Integer \\
AIME 2024 & Competition mathematics & Integer \\
CommonsenseQA & Commonsense reasoning & Multi-choice \\
HotpotQA & Multi-hop QA & Answer span \\
2Wiki\allowbreak MultiHopQA & Multi-hop QA & Answer span \\
\bottomrule
\end{tabular}
\vspace{2mm}
\caption{Composition of BrailleBench. Each data sample is aligned with English, Grade~1 Braille, and Grade~2 Braille formats.}
\label{tab:dataset-stats}
\end{table}

After determining the source tasks, the next step is to define the evaluation protocol in a Braille environment.
Suppose $EN$, $G_1$, and $G_2$ denote English, Grade~1 Braille, and Grade~2 Braille, respectively; the benchmark can contain different input--output settings. Specifically, $G_i{\rightarrow}EN$ tests whether the model can understand and solve a Braille-authored problem. Conversely, $EN{\rightarrow}G_i$ tests whether the model can produce an answer decoded in Braille without first-hand Braille input.
Besides, $G_i{\rightarrow}G_i$ evaluate the end-to-end Braille interaction. In all configurations, the source input, target answer, and task evaluation metrics are consistent with the original $EN{\rightarrow}EN$ setting, which serves as the reference in terms of the capability available to each evaluated model.


In addition, Grade~1 and Grade~2 Braille will result in different difficulty for LLMs to comprehend because of their express rule. 
In Braille ASCII, each printable ASCII character represents one six-dot Braille cell, the basic $2\times3$ dot-pattern unit of Braille.
Thus, Grade~1 Braille yields highly informative cell sequences that maintain closer local correspondence with standard print orthography; within Braille ASCII, this may preserve lexical surface cues observed during pre-training. Conversely, Grade~2 achieves greater cell-level density, though its contractions overload individual symbols with broader semantic meaning, rendering interpretation highly contingent on positional constraints and local context.
Furthermore, fewer cells do not necessarily imply fewer model tokens because segmentation of symbol-heavy strings is model-dependent. 
Then, the next question is how to construct Braille data instances based on the source instances.

\subsection{Braille Data Instance Construction}
\label{sec:construction}
To construct Braille data instances from the English-based ones is non-trivial, since it is not a field-wise character substitution as translation task.
For example, mathematical tasks contain LaTeX-based expressions whose visual structure must be made explicit before transcription. 
Besides, in natural language-based QA datasets, a dropped relation, reordered choice, or unresolved markup fragment can change the original semantics, where UEB indicators and Grade~2 contractions can then amplify that error during transcription. Thus, a task-aware pipeline is required to separate semantic normalization from Braille encoding.

Given a source instance $x$, the pipeline constructs aligned representations $\{x^{EN},x^{G_1},x^{G_2}\}$ in three stages: task-aware normalization, deterministic UEB transcription, and structured record assembly, which is described in the following. 
The same transformation is applied to every task-related field, including questions, answer choices, reference answers, alternative gold answers, and available solutions. Note that no LLM-generated Braille is used as context or ground truth.

\subsubsection{Task-Aware Normalization}
For mathematical data, the pipeline deterministically linearizes common and LaTeX-based expressions into explicit plain text before transcription. Fractions, roots, superscripts, subscripts, relations, Greek letters, and answer boxes are converted while preserving their scope. For example, \texttt{\textbackslash frac\{3\}\{4\}} would become \texttt{(3)/(4)}. 
Presentation-only commands are removed, whereas unsupported commands remain visible as auditable residuals rather than being silently discarded. 
For example, \texttt{x \textbackslash overset\{?\}\{=\} y} remains visible in the normalized text, while \texttt{\textbackslash overset} is recorded as a residual command for inspection or strict-mode rejection.
For natural language data in QA tasks, dataset-specific adapters would map heterogeneous records into a common schema while preserving the question, labeled choices, multiple gold-answer aliases, and solution fields.

\subsubsection{Deterministic UEB Transcription}
We transcribe each normalized field with \texttt{liblouis} version~3.32.0~\cite{liblouis}, using the UEB tables \texttt{en-ueb-g1.ctb} and \texttt{en-ueb-g2.ctb}. 
We use a fixed translation engine, version, and pair of UEB tables to make the transcription deterministic and reproducible while eliminating model-dependent variation. 
Each Braille field is generated from its aligned normalized English field, and the original source field is not overwritten. In this way, the instance- and field-level correspondence across the English, Grade~1, and Grade~2 representations.

\subsubsection{Braille Toolkit}

We further release \textbf{Braille Toolkit}, a benchmark-independent, open-source framework that generalizes this construction methodology beyond BrailleBench. Rather than serving as a thin wrapper around \texttt{liblouis}, it provides reusable task-aware adapters for researchers' own mathematical and QA records, performs auditable normalization, transcribes each field in UEB Grade~1 and Grade~2, and emits synchronized Braille ASCII, Unicode Braille, and dot-number representations. Every converted record retains its original and normalized text, applied rules, residual markup, fallback status, and generated variants. The toolkit also provides fail-closed environment checks, back-translation, cross-format cell validation, and a model-output validator that preserves legal Braille ASCII punctuation, distinguishes print numerals from UEB numerals, and handles edge symbols that resemble Markdown syntax. It therefore supports the construction and evaluation of new UEB datasets without depending on BrailleBench data or model infrastructure. Its mathematical output is linearized content transcribed in English UEB rather than Nemeth Code.

\subsection{Quality Control and Additional Representations}
\label{sec:quality-control}

For BrailleBench, we apply the toolkit's quality controls at the environment, representation, and record levels. We first verify that the specified \texttt{liblouis} tables are available and reproduce known translations.
We then back-translate each normalized field with the grade-matched table as a diagnostic. Since UEB back-translation can be context-sensitive, particularly for isolated mathematical symbols and Grade~2 contractions, disagreement is flagged for inspection rather than automatically treated as evidence of faulty forward transcription. The retained source text, normalization trace, residual markup, and fallback status make each flagged record auditable.

We additionally verify cell-level consistency across Braille ASCII, Unicode Braille patterns, and dot-number notation: all three serializations must resolve to the same ordered Braille cells. Human experts review the generated transcriptions for fidelity and consistency after these automated checks. Together, deterministic generation, cross-representation validation, back-translation, and expert review prevent representation artifacts from being mistaken for model failures.

Whenever Braille appears in the seven core settings, it is represented as Braille ASCII. The release additionally includes the aligned Unicode and dot-number forms generated from the same cells, enabling controlled evaluation without rebuilding the source data. Section~\ref{sec:format-ablation} uses these forms in a surface-format ablation that changes only the serialization while holding the grade, task content, requested answer, and scorer fixed. These additional settings test whether a failure reflects UEB processing more broadly or sensitivity to a particular model-facing representation.

\subsection{Benchmarking on LLMs}
\label{sec:benchmarking}

After construction and validation, a task-specific prompt renderer instantiates each source item under the seven settings. In the core benchmark, task and output instructions remain in English. For Braille-input conditions, only the question and, where applicable, answer choices are replaced by their grade-matched Braille ASCII forms. For Braille-output conditions, the prompt requests an answer in the corresponding grade. Models receive no demonstrations, decoding keys, or in-context translation examples, so paired settings differ only in the task-bearing representation and requested output. \\

\noindent \textbf{Task Adaptation.} The prompt preserves the response convention of each source task. GSM8K requests step-by-step problem solving and a final numerical answer in \texttt{\textbackslash boxed\{\}}; AIME additionally preserves the requirement that the answer be an integer from 0 to 999. CommonsenseQA presents all labeled choices and requests only the selected answer text, while HotpotQA and 2WikiMultiHopQA request a short phrase or name. QA answers follow the marker ``The answer is:''. FULLBR-$G_1$ and FULLBR-$G_2$, which also transcribe the English instruction while retaining English output, are evaluated separately as a prompt ablation rather than added to the core matrix. \\


\noindent \textbf{Braille-Output Validation.} Raw English normalization cannot be applied directly to Braille ASCII because characters such as \texttt{\#}, \texttt{>}, and \texttt{;} are meaningful Braille cells. We first map Unicode or dot-number output, when present, to the corresponding Braille ASCII cells. A format-aware gate then identifies clear print-English responses; mathematical outputs additionally distinguish literal print numerals from valid UEB numerals. Remaining outputs are back-translated with the grade-matched \texttt{liblouis} table and evaluated against the English reference using the original task metric. Empty responses, refusals, and completed but unparseable answers receive zero credit. All models are audited with this same finalized extraction and scoring pipeline.

\section{Experiments}
\label{sec:experiments}

\subsection{Experimental Setup}
\label{sec:experimental-setup}
We conduct a series of comprehensive experiments to address multiple research questions. The details of the experimental setup are depicted as follows. \\

\noindent \textbf{Evaluation Metric and Datasets.}  
The final answer is extracted using task-aware markers, including the last \texttt{\textbackslash boxed\{\}} expression for mathematics and the explicit QA answer marker. GSM8K and AIME~2024 are scored with Math-Verify~\cite{mathverify}, following the general mathematical evaluation protocol of OpenCompass~\cite{cao2026opencompass}. CommonsenseQA (CommQA), HotpotQA, and 2WikiMultiHopQA (2Wiki) use Exact Match (EM) and token-level F1 under FlashRAG normalization~\cite{jin2024flashrag}, taking the maximum score over valid answer aliases. 
The main experiments with EM as the reported metric for QA might be too restricted in Braille scenarios, and we thus also provide F1 and sub-EM as evaluation metrics to analyze whether evaluation bias might exist.
For each non-English condition, \emph{Avg. Drop} is the unweighted mean of the model's $EN{\rightarrow}EN$ score minus its score in that condition across the five datasets. \\

\noindent \textbf{Models.} We evaluate six representative language models, including proprietary ones (Claude Opus~4.8 and Claude Haiku~4.5) and open-weight ones (Llama~3.3~70B Instruct~\cite{metaLlama33}, Llama~3.1~8B Instruct~\cite{grattafiori2024llama3}, and Qwen3~1.7B/32B~\cite{yang2025qwen3}). \\

\noindent\textbf{Inference Protocol.} All evaluations are closed-book, i.e., the models will not receive any demonstrations, retrieved context, or Braille decoding key. 
The maximum generation length is 4,096 tokens for AIME~2024 and 1,024 tokens for the other datasets, following general settings. 
Prompt construction, answer extraction, task metrics, and Braille-output validation follow the benchmark protocol described in Section~\ref{sec:benchmarking}.

\subsection{RQ1: Can LLMs Understand Braille?}

\begin{table*}[t]
\centering
\scriptsize
\renewcommand{\arraystretch}{1.05}
\newcommand{\scoreup}[2]{#1\raisebox{-0.55ex}{\hspace{0.12em}{\fontsize{4.8}{5}\selectfont\textcolor{green!50!black}{$\uparrow$#2}}}}
\newcommand{\scoredown}[2]{#1\raisebox{-0.55ex}{\hspace{0.12em}{\fontsize{4.8}{5}\selectfont\textcolor{red!70!black}{$\downarrow$#2}}}}
\resizebox{\textwidth}{!}{%
\begin{tabular}{
l
*{5}{c c @{\hspace{0.8pt}{\color{gray!55}\vrule width 0.25pt}\hspace{0.8pt}}}
c c
}
\toprule
\textbf{Model}
& \multicolumn{2}{c}{\textbf{GSM8K}}
& \multicolumn{2}{c}{\textbf{AIME 2024}}
& \multicolumn{2}{c}{\textbf{CommonsenseQA}}
& \multicolumn{2}{c}{\textbf{HotpotQA}}
& \multicolumn{2}{c}{\textbf{2Wiki}}
& \multicolumn{2}{c}{\textbf{Avg. Drop}} \\
\cmidrule(lr){2-3}\cmidrule(lr){4-5}\cmidrule(lr){6-7}\cmidrule(lr){8-9}\cmidrule(lr){10-11}\cmidrule(lr){12-13}
& \textbf{G1} & \textbf{G2}
& \textbf{G1} & \textbf{G2}
& \textbf{G1} & \textbf{G2}
& \textbf{G1} & \textbf{G2}
& \textbf{G1} & \textbf{G2}
& \textbf{G1} & \textbf{G2} \\
\midrule
\rowcolor{gray!12}
\multicolumn{13}{c}{\textbf{$E\!\rightarrow\!E$ Reference}} \\
\midrule
Opus 4.8 + CoT & \multicolumn{2}{c}{\textbf{97.4}} & \multicolumn{2}{c}{93.3} & \multicolumn{2}{c}{\textbf{88.8}} & \multicolumn{2}{c}{\textbf{42.7}} & \multicolumn{2}{c}{\textbf{56.1}} & \multicolumn{2}{c}{--} \\
Claude Opus 4.8 & \multicolumn{2}{c}{97.3} & \multicolumn{2}{c}{\textbf{100.0}} & \multicolumn{2}{c}{88.6} & \multicolumn{2}{c}{40.0} & \multicolumn{2}{c}{46.7} & \multicolumn{2}{c}{--} \\
Claude Haiku 4.5 & \multicolumn{2}{c}{96.2} & \multicolumn{2}{c}{53.3} & \multicolumn{2}{c}{81.7} & \multicolumn{2}{c}{26.4} & \multicolumn{2}{c}{20.9} & \multicolumn{2}{c}{--} \\
Llama 3.3 70B & \multicolumn{2}{c}{95.1} & \multicolumn{2}{c}{23.3} & \multicolumn{2}{c}{80.4} & \multicolumn{2}{c}{29.7} & \multicolumn{2}{c}{29.7} & \multicolumn{2}{c}{--} \\
Llama 3.1 8B & \multicolumn{2}{c}{83.9} & \multicolumn{2}{c}{6.7} & \multicolumn{2}{c}{43.4} & \multicolumn{2}{c}{21.4} & \multicolumn{2}{c}{25.0} & \multicolumn{2}{c}{--} \\
Qwen3 32B & \multicolumn{2}{c}{94.7} & \multicolumn{2}{c}{20.0} & \multicolumn{2}{c}{83.5} & \multicolumn{2}{c}{21.4} & \multicolumn{2}{c}{28.3} & \multicolumn{2}{c}{--} \\
Qwen3 1.7B & \multicolumn{2}{c}{69.4} & \multicolumn{2}{c}{10.0} & \multicolumn{2}{c}{37.4} & \multicolumn{2}{c}{12.7} & \multicolumn{2}{c}{18.3} & \multicolumn{2}{c}{--} \\
\midrule
\rowcolor{gray!12}
\multicolumn{13}{c}{\textbf{$E\!\rightarrow\!G$}} \\
\midrule
Claude Opus 4.8 & \textbf{\scoredown{91.7}{5.6}} & \textbf{\scoredown{86.0}{11.3}} & \textbf{\scoredown{43.3}{56.7}} & \textbf{\scoredown{53.3}{46.7}} & \scoredown{55.0}{33.6} & \scoredown{46.5}{42.1} & \textbf{\scoredown{28.5}{11.5}} & \textbf{\scoredown{25.3}{14.7}} & \textbf{\scoredown{40.0}{6.7}} & \textbf{\scoredown{27.3}{19.4}} & 22.8 & 26.8 \\
Claude Haiku 4.5 & \scoredown{5.0}{91.2} & \scoredown{0.1}{96.1} & \scoredown{0.0}{53.3} & \scoredown{0.0}{53.3} & \textbf{\scoredown{75.3}{6.4}} & \textbf{\scoredown{72.5}{9.2}} & \scoredown{11.3}{15.1} & \scoredown{12.9}{13.5} & \scoredown{8.9}{12.0} & \scoredown{11.0}{9.9} & 35.6 & 36.4 \\
Llama 3.3 70B & \scoredown{4.9}{90.2} & \scoredown{0.0}{95.1} & \scoredown{3.3}{20.0} & \scoredown{0.0}{23.3} & \scoredown{50.0}{30.4} & \scoredown{29.7}{50.7} & \scoredown{11.1}{18.6} & \scoredown{6.2}{23.5} & \scoredown{12.1}{17.6} & \scoredown{3.8}{25.9} & 35.4 & 43.7 \\
Llama 3.1 8B & \scoredown{2.6}{81.3} & \scoredown{0.0}{83.9} & \scoredown{0.0}{6.7} & \scoredown{0.0}{6.7} & \scoredown{17.2}{26.2} & \scoredown{17.4}{26.0} & \scoredown{1.0}{20.4} & \scoredown{0.9}{20.5} & \scoredown{0.4}{24.6} & \scoredown{0.5}{24.5} & 31.8 & 32.3 \\
Qwen3 32B & \scoredown{6.1}{88.6} & \scoredown{2.0}{92.7} & \scoredown{3.3}{16.7} & \scoredown{3.3}{16.7} & \scoredown{57.8}{25.7} & \scoredown{43.0}{40.5} & \scoredown{2.5}{18.9} & \scoredown{2.7}{18.7} & \scoredown{2.2}{26.1} & \scoredown{0.9}{27.4} & 35.2 & 39.2 \\
Qwen3 1.7B & \scoredown{2.5}{66.9} & \scoredown{0.0}{69.4} & \scoredown{0.0}{10.0} & \scoredown{0.0}{10.0} & \scoredown{22.2}{15.2} & \scoredown{26.5}{10.9} & \scoredown{0.1}{12.6} & \scoredown{0.1}{12.6} & \scoredown{0.0}{18.3} & \scoredown{0.0}{18.3} & 24.6 & 24.2 \\
\midrule
\rowcolor{gray!12}
\multicolumn{13}{c}{\textbf{$G\!\rightarrow\!E$}} \\
\midrule
Claude Opus 4.8 & \textbf{\scoredown{89.8}{7.5}} & \textbf{\scoredown{67.9}{29.4}} & \textbf{\scoredown{90.0}{10.0}} & \textbf{\scoredown{76.7}{23.3}} & \textbf{\scoredown{86.9}{1.7}} & \textbf{\scoredown{34.6}{54.0}} & \textbf{\scoredown{38.8}{1.2}} & \textbf{\scoredown{12.0}{28.0}} & \textbf{\scoreup{49.5}{2.9}} & \textbf{\scoredown{6.9}{39.8}} & 3.5 & 34.9 \\
Claude Haiku 4.5 & \scoredown{35.6}{60.6} & \scoredown{11.7}{84.5} & \scoredown{3.3}{50.0} & \scoredown{3.3}{50.0} & \scoredown{30.9}{50.8} & \scoredown{24.0}{57.7} & \scoredown{25.6}{0.8} & \scoredown{10.3}{16.1} & \scoreup{23.9}{3.0} & \scoredown{7.2}{13.7} & 31.8 & 44.4 \\
Llama 3.3 70B & \scoredown{7.3}{87.8} & \scoredown{3.7}{91.4} & \scoredown{0.0}{23.3} & \scoredown{0.0}{23.3} & \scoredown{8.9}{71.5} & \scoredown{4.1}{76.3} & \scoredown{18.6}{11.1} & \scoredown{3.0}{26.7} & \scoredown{17.3}{12.4} & \scoredown{0.8}{28.9} & 41.2 & 49.3 \\
Llama 3.1 8B & \scoredown{4.6}{79.3} & \scoredown{1.4}{82.5} & \scoredown{0.0}{6.7} & \scoredown{0.0}{6.7} & \scoredown{0.2}{43.2} & \scoredown{0.5}{42.9} & \scoredown{10.9}{10.5} & \scoredown{0.7}{20.7} & \scoredown{7.1}{17.9} & \scoredown{0.5}{24.5} & 31.5 & 35.5 \\
Qwen3 32B & \scoredown{13.0}{81.7} & \scoredown{7.7}{87.0} & \scoredown{0.0}{20.0} & \scoredown{3.3}{16.7} & \scoredown{57.1}{26.4} & \scoredown{17.4}{66.1} & \scoredown{14.2}{7.2} & \scoredown{1.3}{20.1} & \scoredown{20.3}{8.0} & \scoredown{0.9}{27.4} & 28.7 & 43.5 \\
Qwen3 1.7B & \scoredown{2.4}{67.0} & \scoredown{1.4}{68.0} & \scoredown{0.0}{10.0} & \scoredown{0.0}{10.0} & \scoredown{9.6}{27.8} & \scoredown{1.5}{35.9} & \scoredown{7.1}{5.6} & \scoredown{0.7}{12.0} & \scoredown{8.3}{10.0} & \scoredown{0.5}{17.8} & 24.1 & 28.7 \\
\midrule
\rowcolor{gray!12}
\multicolumn{13}{c}{\textbf{$G\!\rightarrow\!G$}} \\
\midrule
Claude Opus 4.8 & \textbf{\scoredown{86.3}{11.0}} & \textbf{\scoredown{63.5}{33.8}} & \textbf{\scoredown{73.3}{26.7}} & \textbf{\scoredown{50.0}{50.0}} & \textbf{\scoredown{23.3}{65.3}} & \textbf{\scoredown{14.4}{74.2}} & \textbf{\scoreup{42.3}{2.3}} & \textbf{\scoredown{9.4}{30.6}} & \textbf{\scoreup{57.8}{11.1}} & \textbf{\scoredown{6.5}{40.2}} & 17.9 & 45.8 \\
Claude Haiku 4.5 & \scoredown{14.5}{81.7} & \scoredown{7.1}{89.1} & \scoredown{0.0}{53.3} & \scoredown{0.0}{53.3} & \scoredown{15.2}{66.5} & \scoredown{8.6}{73.1} & \scoredown{21.1}{5.3} & \scoredown{4.9}{21.5} & \scoreup{21.6}{0.7} & \scoredown{4.9}{16.0} & 41.2 & 50.6 \\
Llama 3.3 70B & \scoredown{1.8}{93.3} & \scoredown{0.2}{94.9} & \scoredown{0.0}{23.3} & \scoredown{0.0}{23.3} & \scoredown{10.6}{69.8} & \scoredown{4.3}{76.1} & \scoredown{17.8}{11.9} & \scoredown{1.9}{27.8} & \scoredown{19.7}{10.0} & \scoredown{1.1}{28.6} & 41.7 & 50.1 \\
Llama 3.1 8B & \scoredown{0.5}{83.4} & \scoredown{0.0}{83.9} & \scoredown{0.0}{6.7} & \scoredown{0.0}{6.7} & \scoredown{0.1}{43.3} & \scoredown{2.5}{40.9} & \scoredown{4.1}{17.3} & \scoredown{0.1}{21.3} & \scoredown{4.5}{20.5} & \scoredown{0.0}{25.0} & 34.2 & 35.6 \\
Qwen3 32B & \scoredown{3.3}{91.4} & \scoredown{1.6}{93.1} & \scoredown{0.0}{20.0} & \scoredown{0.0}{20.0} & \scoredown{9.3}{74.2} & \scoredown{1.1}{82.4} & \scoredown{10.9}{10.5} & \scoredown{1.4}{20.0} & \scoredown{20.3}{8.0} & \scoredown{3.5}{24.8} & 40.8 & 48.1 \\
Qwen3 1.7B & \scoredown{0.0}{69.4} & \scoredown{0.3}{69.1} & \scoredown{0.0}{10.0} & \scoredown{0.0}{10.0} & \scoredown{2.5}{34.9} & \scoredown{0.2}{37.2} & \scoredown{1.5}{11.2} & \scoredown{0.1}{12.6} & \scoredown{1.3}{17.0} & \scoredown{0.0}{18.3} & 28.5 & 29.4 \\
\bottomrule
\end{tabular}%
}
\vspace{2mm}
\caption{Main experimental results organized by input--output direction. All $\Delta$ annotations continue to use the corresponding model's standard EN-EN score. Bold denotes the best task score within each directional setting and grade.}
\label{tab:core-results}
\end{table*}


\noindent\textbf{Across Language Settings and Models.}
As shown in Table~\ref{tab:core-results}, introducing Braille leads to a broad performance decline across models and datasets. The strongest English models generally retain more capability in Braille, with Opus showing the most consistent performance, while smaller models tend to lose most of their original task capability. However, the ordering is not strictly determined by model scale: several models exhibit isolated strengths on particular QA tasks but fail to transfer them across task families. Mathematics shows the clearest separation between models, whereas QA performance is more variable and includes a small number of localized improvements that we examine separately.

The three interaction directions further reveal that Braille understanding is not a single capability. Models may partially recover a Braille-authored question under $G\!\rightarrow\!E$ yet fail to express an already known answer under $E\!\rightarrow\!G$. Requiring both abilities under $G\!\rightarrow\!G$ generally compounds these failures and produces the least reliable interaction. Across all three directions, Grade~2 is more challenging than Grade~1, particularly when contracted Braille must be interpreted on the input side. Overall, the results show that strong English task performance does not automatically transfer to Braille reading, accessible expression, or end-to-end interaction. Table~\ref{tab:prompt-examples} shows complete prompts for representative GSM8K and CommonsenseQA instances. \\


\begin{table*}[t]
\centering
\scriptsize
\renewcommand{\arraystretch}{1.15}
\begin{tabular}{>{\raggedright\arraybackslash}p{0.11\textwidth} >{\raggedright\arraybackslash}p{0.76\textwidth} >{\raggedright\arraybackslash}p{0.07\textwidth}}
\toprule
\textbf{Variant} & \textbf{Model-Facing Prompt} & \textbf{Gold} \\
\midrule
\multicolumn{3}{c}{\textbf{GSM8K}} \\
\midrule
EN-EN & \texttt{Solve the following math problem step by step. Answer in plain English. Put your final numerical answer within \textbackslash boxed\{\}. Problem: Janet's ducks lay 16 eggs per day. She eats three for breakfast every morning and bakes muffins for her friends every day with four. She sells the remainder at the farmers' market daily for \$2 per fresh duck egg. How much in dollars does she make every day at the farmers' market?} & \texttt{18} \\
\addlinespace[2pt]
G1-EN & \texttt{The following question is written in Grade 1 (uncontracted) Braille ASCII notation. Solve the following math problem step by step. Answer in plain English. Put your final numerical answer within \textbackslash boxed\{\}. Problem: ,janet's ducks lay \#af eggs per day4 ,she eats three for breakfast every morning and bakes muffins for her friends every day with four4 ,she sells the remainder at the farmers' market daily for `s\#b per fresh duck egg4 ,how much in dollars does she make every day at the farmers' market8} & \texttt{18} \\
\addlinespace[2pt]
G2-EN & \texttt{The following question is written in Grade 2 (contracted) Braille ASCII notation. Solve the following math problem step by step. Answer in plain English. Put your final numerical answer within \textbackslash boxed\{\}. Problem: ,janet's ducks lay \#af e7s p\} "d4 ,\%e eats ?ree = br1kfa/ e morn+ \& bakes mu69s = h\} frs e "d ) f|r4 ,\%e sells ! rema9d\} at ! f>m\}s' m>ket daily = `s\#b p\} fre\% duck egg4 ,h\{ m* 9 doll>s does \%e make e "d at ! f>m\}s' m>ket8} & \texttt{18} \\
\addlinespace[2pt]
FULLBR-G1 & \texttt{,solve the following math problem step by step4 ,answer in plain ,english4 ,put your final numerical answer within \_*boxed\_<\_>4 ,problem3 ,janet's ducks lay \#af eggs per day4 ,she eats three for breakfast every morning and bakes muffins for her friends every day with four4 ,she sells the remainder at the farmers' market daily for `s\#b per fresh duck egg4 ,how much in dollars does she make every day at the farmers' market8} & \texttt{18} \\
\addlinespace[2pt]
FULLBR-G2 & \texttt{,solve ! foll\{+ ma? problem /ep by /ep4 ,answ\} 9 pla9 ,5gli\%4 ,put yr f9al num\}ical answ\} )9 \_*box\$\_<\_>4 ,problem3 ,janet's ducks lay \#af e7s p\} "d4 ,\%e eats ?ree = br1kfa/ e morn+ \& bakes mu69s = h\} frs e "d ) f|r4 ,\%e sells ! rema9d\} at ! f>m\}s' m>ket daily = `s\#b p\} fre\% duck egg4 ,h\{ m* 9 doll>s does \%e make e "d at ! f>m\}s' m>ket8} & \texttt{18} \\
\midrule
\multicolumn{3}{c}{\textbf{CommonsenseQA}} \\
\midrule
EN-EN & \texttt{Answer the following multiple-choice question. Answer in plain English. Write only the answer text after 'The answer is: '. Question: A revolving door is convenient for two direction travel, but it also serves as a security measure at a what? Choices: A) bank B) library C) department store D) mall E) new york} & \texttt{bank} \\
\addlinespace[2pt]
G1-EN & \texttt{The following question is written in Grade 1 (uncontracted) Braille ASCII notation. Answer the following multiple-choice question. Answer in plain English. Write only the answer text after 'The answer is: '. Question: ,a revolving door is convenient for two direction travel1 but it also serves as a security measure at a what8 Choices: ,a"> bank ,b"> library ,c"> department store ,d"> mall ,e"> new york} & \texttt{bank} \\
\addlinespace[2pt]
G2-EN & \texttt{The following question is written in Grade 2 (contracted) Braille ASCII notation. Answer the following multiple-choice question. Answer in plain English. Write only the answer text after 'The answer is: '. Question: ,a revolv+ door is 3v5i5t = two direc;n travel1 b x al s\}ves z a secur;y m1sure at a :at8 Choices: ,a"> bank ;,b"> libr>y ;,c"> de"p;t /ore ;,d"> mall ;,e"> new york} & \texttt{bank} \\
\addlinespace[2pt]
FULLBR-G1 & \texttt{,answer the following multiple-choice question4 ,answer in plain ,english4 ,write only the answer text after ',the answer is3 '4 ,question3 ,a revolving door is convenient for two direction travel1 but it also serves as a security measure at a what8 ,choices3 ,a"> bank ,b"> library ,c"> department store ,d"> mall ,e"> new york} & \texttt{bank} \\
\addlinespace[2pt]
FULLBR-G2 & \texttt{,answ\} ! foll\{+ multiple-*oice "q4 ,answ\} 9 pla9 ,5gli\%4 ,write only ! answ\} text af ',! answ\} is3 '4 ,"q3 ,a revolv+ door is 3v5i5t = two direc;n travel1 b x al s\}ves z a secur;y m1sure at a :at8 ,*oices3 ,a"> bank ;,b"> libr>y ;,c"> de"p;t /ore ;,d"> mall ;,e"> new york} & \texttt{bank} \\
\bottomrule
\end{tabular}
\vspace{2mm}
\caption{Representative GSM8K and CommonsenseQA prompt variants. FULLBR-G1 and FULLBR-G2 translate both the instruction and task content into Braille, retain English as the requested answer language, and are evaluated separately from the seven core configurations. Line breaks are collapsed for compact presentation.}
\label{tab:prompt-examples}
\end{table*}

\noindent\textbf{Performance in Grade~1 versus Grade~2.}
Across models and datasets, Grade~2 is consistently more difficult than Grade~1 when Braille must be read from the input. This difference is also reflected in end-to-end interaction, where the loss introduced by Grade~2 compounds with the requirement to generate a Braille answer. By contrast, the grade difference is smaller and less consistent when models only need to produce Braille from an English question. The gap is most visible on language-intensive QA tasks, while some mathematical settings exhibit a floor effect because several models already perform near zero under Grade~1. Overall, the results localize the main Grade~2 difficulty to accessing contracted input rather than simply producing a shorter Braille output.

This difficulty cannot be explained by sequence length alone. Grade~2 uses fewer cells, but its contractions assign more lexical information to individual cells and make their interpretation dependent on position and context. As shown in Tables~\ref{tab:response-overhead} and Table~\ref{tab:length-ratio-all}, models that attempt to process Grade~2 often produce longer decoding traces, incomplete answers, or empty responses despite receiving a shorter Braille sequence. Grade~2 therefore trades surface length for contextual ambiguity: local contraction errors can alter words, entities, and relations before task completion begins, and these errors become increasingly consequential as the input grows longer or requires multi-step understanding. \\

\begin{table}[t]
\centering
\scriptsize
\resizebox{\columnwidth}{!}{%
\begin{tabular}{@{}l r r r r r@{}}
\toprule
\textbf{Dataset} & \textbf{EN} & \textbf{G1} & \textbf{G2} & \shortstack{\textbf{G2$\rightarrow$E}\\\textbf{Empty}} & \shortstack{\textbf{G2$\rightarrow$G2}\\\textbf{Empty}} \\
\midrule
GSM8K & 253 & 978 & 1,518 & 12.1 & 10.4 \\
AIME 2024 & 1,289 & 2,214 & 2,555 & 20.0 & 23.3 \\
CommonsenseQA & 25 & 29 & 519 & 54.7 & 56.9 \\
HotpotQA & 88 & 516 & 910 & 64.9 & 72.2 \\
2Wiki & 163 & 588 & 940 & 71.6 & 78.4 \\
\bottomrule
\end{tabular}
}
\vspace{2mm}
\caption{The average response length in non-empty examples for Opus under EN-EN, G1-EN, and G2-EN language settings. The right two columns report the percentage of empty output examples in G2-EN and G2-G2.}
\label{tab:response-overhead}
\end{table}

\begin{table}[t]
\centering
\resizebox{\columnwidth}{!}{%
\begin{tabular}{@{}l r r r r r@{}}
\toprule
\textbf{Model} & \textbf{GSM8K} & \textbf{AIME} & \textbf{CSQA} & \textbf{Hotpot} & \textbf{2Wiki} \\
\midrule
Claude Opus 4.8 & 6.0 & 2.0 & 20.5 & 10.4 & 5.8 \\
Claude Haiku 4.5 & 2.2 & 1.0 & 17.8 & 4.3 & 3.1 \\
Llama 3.3 70B & 2.3 & 0.7 & 0.7 & 1.1 & 1.1 \\
Llama 3.1 8B & 1.7 & 1.2 & 3.8 & 4.0 & 4.2 \\
Qwen3 32B & 1.5 & 1.1 & 0.8 & 2.2 & 1.6 \\
\bottomrule
\end{tabular}
}
\vspace{2mm}
\caption{Ratio of average non-empty G2-EN response length to the same model's EN-EN response length. Values below one indicate that the model produced less text under Braille input rather than externalizing a decode.}
\label{tab:length-ratio-all}
\end{table}

\noindent\textbf{Counterintuitive Results in Multi-hop QA Tasks.}
\label{sec:2wiki-behavior}
Intuitively, the English-English reference should be the upper bound for the results under Braille conditions.
However, as shown in Table~\ref{tab:core-results}, the Claude models exhibit performance improvement from Braille (Grade 1) to English and to Braille settings on two multi-hop QA datasets.

We investigate the potential reasons, which are mainly attributed to two aspects.
On one hand, as shown in Table~\ref{tab:2wiki-reasoning}, the Opus model tend response the question directly rather than conduct more steps of reasoning in the EN-EN setting, which is more ``aggressive'' compared to the setting with Grade 1 Braille. Thus, the model might not be able to produce a correct answer at once on this more challenging multi-hop QA dataset, while the model becomes more careful when involving Braille and thus has the higher chance to arrive at the correct answer, although with longer reasoning chains.
On the other hand, Table~\ref{tab:2wiki-reasoning} compares strict EM with Sub-EM and F1 among different languages and CoT settings.
We can see the potential bias in terms of the evaluation metrics. The manually activated CoT can further improve the results across all metrics, which confirms our previous conjecture. In addition, A more relaxed evaluation metric, i.e., Sub-EM and F1, can better reflect the performance gap between the intuitive upper bound and the results in Braille.

\begin{table}[t]
\centering
\begin{tabular}{l r r r}
\toprule
\textbf{Configuration} & \shortstack{\textbf{Reasoning}\\\textbf{Activation}} & \shortstack{\textbf{Reasoning}\\\textbf{Deactivation}} & \shortstack{\textbf{Average}\\\textbf{Length}} \\
\midrule
EN-EN + CoT & 99\% & 1\% & 278 \\
EN-EN & 57\% & 43\% & 163 \\
G1-EN & 97\% & 3\% & 588 \\
G1-G1 & 99\% & 1\% & 790 \\
\midrule
\textbf{Configuration} & \textbf{EM} & \textbf{Sub-EM} & \textbf{F1} \\
\midrule
EN-EN + CoT & 56.1 & 64.0 & 65.5 \\
EN-EN & 46.7 & 49.9 & 52.3 \\
G1-EN & 49.5 & 58.7 & 59.5 \\
G1-G1 & 57.8 & 60.1 & 64.1 \\
\bottomrule
\end{tabular}
\vspace{2mm}
\caption{Response activation and length of Opus on 2WikiMultiHopQA.}
\label{tab:2wiki-reasoning}
\end{table}

\subsection{RQ2: Can LLMs Understand a Pure Braille Request?}
\label{sec:diagnostics}

In the Braille-input settings, the user question is written in Braille, while the task instruction remains in English to help the model identify the task and expected answer format, which might provide biased assistance. To test whether LLMs can understand a fully Braille input without any English instruction, we translate both the instruction and the problem into Grade~1 (FULLBR-G1) or Grade~2 (FULLBR-G2) Braille, respectively. 

\begin{table}
\centering
\scriptsize
\begin{tabular}{l c c c c c}
\toprule
\textbf{Model} & \textbf{GSM8K} & \textbf{AIME} & \textbf{CommQA} & \textbf{HotpotQA} & \textbf{2Wiki} \\
\midrule
\multicolumn{6}{c}{\textbf{G1-EN / FULLBR-G1}} \\
\midrule
Opus 4.8 & 89 / 73 & 90 / 90 & 86 / 87 & 43 / 40 & 53 / 46 \\
Haiku 4.5 & 34 / 5 & 3.3 / 6.7 & 28 / 14 & 36 / 21 & 34 / 19 \\
Llama 3.3 70B & 9 / 5 & 0 / 0 & 11 / 20 & 18 / 21 & 22 / 14 \\
Llama 3.1 8B & 6 / 7 & 0 / 0 & 0 / 0 & 11 / 12 & 13 / 20 \\
Qwen3 32B & 14 / 4 & 0 / 0 & 54 / 27 & 11 / 16 & 23 / 30 \\
\midrule
\multicolumn{6}{c}{\textbf{G2-EN / FULLBR-G2}} \\
\midrule
Claude Opus 4.8 & 63 / 44 & 76.7 / 0 & 32 / 8 & 13 / 2 & 10 / 3 \\
Claude Haiku 4.5 & 9 / 3 & 3.3 / 3.3 & 20 / 4 & 10 / 3 & 11 / 1 \\
Llama 3.3 70B & 3 / 2 & 0 / 0 & 4 / 6 & 2 / 3 & 0 / 1 \\
Llama 3.1 8B & 1 / 1 & 0 / 0 & 0 / 1 & 0 / 1 & 0 / 0 \\
Qwen3 32B & 7 / 5 & 3.3 / 6.7 & 19 / 4 & 0 / 0 & 1 / 0 \\
\bottomrule
\end{tabular}
\vspace{2mm}
\caption{Effect of transcribing the task instruction into Braille across all five diagnostic subsets. Each cell reports G1-EN / FULLBR-G1 or G2-EN / FULLBR-G2 on the same 100-example (30-example on AIME 2024 prefix.}
\label{tab:fullbr-results}
\end{table}

Table~\ref{tab:fullbr-results} show that transcribing both the instruction and the problem generally lowers performance, and the effect depends strongly on grade and model. Under FULLBR-G1, Opus retains substantial accuracy across all five tasks, while the other models decline from their mixed-prompt results or remain the same; occasional increases occur mainly at low absolute scores. Under FULLBR-G2, performance becomes uniformly weak: even Opus preserves meaningful accuracy only on GSM8K, and the remaining model--task combinations are close to the floor. These results show that the English instruction helps most models understand what task they should perform. Opus can still answer many FULLBR-G1 problems, showing that it can understand some requests written entirely in Grade~1 Braille, but FULLBR-G2 is difficult for every model. We next ask whether the same Braille content becomes easier or harder when it is represented in different digital formats.

\subsection{RQ3: How Expression Form of Braille Affect the Results?}
\label{sec:format-ablation}

To separate knowledge of Braille rules from sensitivity to serialization, we render the same cells as Braille ASCII, Unicode patterns, and dot numbers while holding the grade, task instruction, requested English answer, and scorer fixed. 

\begin{table}
\centering
\scriptsize
\begin{tabular}{l c c c c c}
\toprule
\textbf{Model} & \textbf{GSM8K} & \textbf{AIME} & \textbf{CommQA} & \textbf{HotpotQA} & \textbf{2Wiki} \\
\midrule
\multicolumn{6}{c}{\textbf{Grade 1: ASCII / Unicode / Dots}} \\
\midrule
Opus 4.8 & 89/0/0 & 90/0/0 & 86/0/0 & 43/0/0 & 53/0/0 \\
Haiku 4.5 & 34/48/1 & 3.3/10/0 & 28/19/0 & 36/24/0 & 34/29/0 \\
Llama 3.3 70B & 9/4/0 & 0/0/0 & 11/2/0 & 18/2/0 & 22/0/0 \\
Llama 3.1 8B & 6/0/1 & 0/0/0 & 0/0/0 & 11/0/0 & 13/0/0 \\
Qwen3 32B & 14/19/0 & 0/6.7/0 & 54/23/0 & 11/5/0 & 23/6/0 \\
\midrule
\multicolumn{6}{c}{\textbf{Grade 2: ASCII / Unicode / Dots}} \\
\midrule
Opus 4.8 & 63/0/0 & 76.7/0/0 & 32/0/0 & 13/0/0 & 10/0/0 \\
Haiku 4.5 & 9/30/5 & 3.3/3.3/0 & 20/9/0 & 10/11/0 & 11/11/0 \\
Llama 3.3 70B & 3/0/2 & 0/0/0 & 4/1/0 & 2/0/0 & 0/0/0 \\
Llama 3.1 8B & 1/0/0 & 0/0/0 & 0/0/0 & 0/0/0 & 0/0/0 \\
Qwen3 32B & 7/9/0 & 3.3/3.3/0 & 19/11/0 & 0/0/0 & 1/1/1 \\
\bottomrule
\end{tabular}
\vspace{2mm}
\caption{Surface-format comparison on the diagnostic subsets. Each cell reports Braille ASCII / Unicode Braille / dot-number notation for the indicated grade.}
\label{tab:format-results}
\end{table}

Table~\ref{tab:format-results} shows that identical Braille cells produce markedly different results across surface representations. Braille ASCII gives the strongest and most stable performance across models, grades, and tasks. Unicode improves selected mathematics results for Haiku and Qwen3~32B, but usually underperforms ASCII elsewhere, while Opus scores zero on every evaluated task in this format. Dot notation is almost uniformly at the floor for both grades.

These results show that a digital Braille representation is not a neutral rendering choice for current LLMs: performance depends on the serialized symbols through which the cells are presented. Unicode can be useful for particular model--task pairs, but its gains do not generalize, while expanding cells into dot-number sequences makes the input consistently harder rather than more explicit. Since choosing a more favorable surface form does not remove the broader understanding gap, we next test whether directly supplying the Braille mappings can repair it. i.e., improving the Braille context understanding.

\subsection{RQ4: Can an Explicit Braille Reference Help Context Understanding?}

To understand whether missing Braille mappings from difficulty applying them over a sequence, we prepend a \texttt{liblouis}-verified, grade-matched reference of letters, indicators, punctuation, and common contractions, without worked examples or answers. The Braille problem, English instruction and output, evaluation subset, and scorer remain fixed. 

\begin{table}
\centering
\scriptsize
\begin{tabular}{l c c c c c}
\toprule
\textbf{Model} & \textbf{GSM8K} & \textbf{AIME} & \textbf{CommQA} & \textbf{HotpotQA} & \textbf{2Wiki} \\
\midrule
\multicolumn{6}{c}{\textbf{Grade 1: no reference / with reference}} \\
\midrule
Opus 4.8 & 89.0/97.0 & 90.0/90.0 & 86.0/82.0 & 43.0/45.0 & 53.0/51.0 \\
Haiku 4.5 & 34.0/73.0 & 3.3/23.3 & 28.0/37.0 & 36.0/39.0 & 34.0/37.0 \\
Llama 3.3 70B & 9.0/46.0 & 0.0/6.7 & 11.0/12.0 & 18.0/19.0 & 22.0/20.0 \\
Llama 3.1 8B & 6.0/11.0 & 0.0/0.0 & 0.0/1.0 & 11.0/8.0 & 13.0/12.0 \\
Qwen3 32B & 14.0/47.0 & 0.0/6.7 & 54.0/37.0 & 11.0/13.0 & 23.0/24.0 \\
\midrule
\multicolumn{6}{c}{\textbf{Grade 2: no reference / with reference}} \\
\midrule
Opus 4.8 & 63.0/30.0 & 76.7/6.7 & 32.0/2.0 & 13.0/0.0 & 10.0/1.0 \\
Haiku 4.5 & 9.0/47.0 & 3.3/13.3 & 20.0/14.0 & 10.0/8.0 & 11.0/17.0 \\
Llama 3.3 70B & 3.0/8.0 & 0.0/3.3 & 4.0/11.0 & 2.0/0.0 & 0.0/0.0 \\
Llama 3.1 8B & 1.0/1.0 & 0.0/0.0 & 0.0/3.0 & 0.0/0.0 & 0.0/0.0 \\
Qwen3 32B & 7.0/13.0 & 3.3/0.0 & 19.0/6.0 & 0.0/0.0 & 1.0/0.0 \\
\bottomrule
\end{tabular}
\vspace{2mm}
\caption{Ablation on the effectiveness of completed Braille-reference. Each cell compares the same subset without/with the supplied mapping.} 
\label{tab:cheatsheet-results}
\end{table}

The results are shown in Table~\ref{tab:cheatsheet-results}, where the explicit reference is most effective for Grade~1 mathematics. It substantially improves GSM8K for Haiku, Llama~3.3~70B, and Qwen3~32B, and also raises AIME accuracy for several of these models. The gains are smaller and inconsistent on the QA datasets, where some model--task pairs improve while others decline. Grade~2 is less effective: the reference helps selected weaker-model mathematics results, but offers no consistent benefit across QA and sharply reduces Opus performance.

The contrast suggests that incomplete knowledge of local symbols is one source of Grade~1 failure, especially when recovering numbers and short expressions is sufficient to expose the mathematical problem. A static reference is less useful for longer linguistic inputs and Grade~2 contractions, where the correct expansion depends on lexical position and surrounding context rather than on a one-to-one lookup. Because access to the mapping does not guarantee successful sequence-level decoding, we next isolate where performance begins to break down across characters, numbers, words, and sentences.

\subsection{RQ5: Where Does Braille Understanding Break Down?}

To localize failures before downstream task completion, we remove the task and request direct Braille-to-English transcription of characters, numbers, Grade~1 words, contraction-bearing Grade~2 words, and sentences, yielding 446 items per model across two grades. Table~\ref{tab:probe-results} reports normalized exact match at all five scales and punctuation-insensitive sentence EM as an additional diagnostic.

\begin{table}[h]
\centering
\begin{tabular}{l r r r r c}
\toprule
\textbf{Model} & \textbf{Characters} & \textbf{Numbers} & \textbf{G1} & \textbf{G2} & \textbf{Sentences} \\
\midrule
\multicolumn{6}{c}{\textbf{Grade 1}} \\
\midrule
Opus 4.8 & 97 & 97 & 100 & 98 & 40/88 \\
Haiku 4.5 & 90 & 33 & 88 & 100 & 34/70 \\
Llama 3.3 70B & 92 & 13 & 100 & 98 & 8/56 \\
Llama 3.1 8B & 70 & 10 & 75 & 88 & 18/46 \\
Qwen3 32B & 68 & 37 & 100 & 100 & 30/72 \\
\midrule
\multicolumn{6}{c}{\textbf{Grade 2}} \\
\midrule
Opus 4.8 & 92 & 100 & 90 & 72 & 2/28 \\
Haiku 4.5 & 18 & 33 & 58 & 35 & 0/10 \\
Llama 3.3 70B & 35 & 10 & 70 & 23 & 0/4 \\
Llama 3.1 8B & 82 & 10 & 52 & 9 & 0/4 \\
Qwen3 32B & 13 & 23 & 58 & 16 & 0/4 \\
\bottomrule
\end{tabular}
\vspace{2mm}
\caption{Complete direct-decoding probe results. All cells report strict EM; sentence cells additionally report punctuation-insensitive EM after the slash. Grade~1 and Grade~2 denote the prompt encoding, while the two word columns denote the type of lexical item being transcribed.}
\label{tab:probe-results}
\end{table}

Table~\ref{tab:probe-results} shows a clear gap between recognizing local Grade~1 units and decoding complete sequences. Most models achieve high exact match on Grade~1 characters and words, yet all except Opus perform poorly on numbers, indicating difficulty maintaining the number indicator's effect across subsequent cells. Sentence accuracy is much lower than word accuracy, although removing punctuation recovers a substantial portion of Grade~1 errors. The Grade~2 condition is harder at the lexical and sentence scales: contraction-bearing word accuracy falls, and strict sentence EM is zero for every model except Opus, with only limited recovery when punctuation is ignored.

The scale-wise decline localizes the principal failure before downstream task completion. Current models often possess fragments of Braille knowledge, but number-state tracking, contextual contraction resolution, and the accumulation of local errors prevent that knowledge from supporting reliable sentence-level understanding. These diagnostic results clarify that downstream task failures can arise from sequence-level Braille decoding errors, rather than from insufficient task capability alone.

\section{Conclusion and Future Work}

In this paper, we investigate whether LLM capabilities measured in common print English remain accessible through Braille. To this end, we propose BrailleBench, which contains English, UEB Grade~1, and UEB Grade~2 in different configurations to evaluate Braille comprehension from different aspects.
We use a self-designed toolkit to produce Braille to ensure quality and avoid evaluation bias.
Experiments on six representative LLMs show that English capability does not transfer reliably to Braille. 
Overall, some of the existing proprietary models might have fragments of Braille knowledge but lack robust Braille interaction capability.
All experimental observations provide valuable guidance for the development of future Braille AI systems. 

In existing systems, the Braille-output validator is deliberately conservative but cannot perfectly distinguish lowercase print English from Grade~1 Braille ASCII when the byte strings coincide. Thus, they interpret its English-output rate as a lower bound and rely more heavily on numbers and contraction-bearing outputs to establish deliberate Braille encoding.
Besides, API-hosted models expose neither tokenizer traces nor moderation decisions. Response length and empty-output patterns reveal observable costs and failures, but their internal causes remain hypotheses.
To develop truly inclusive AI systems that can also benefit vulnerable groups, future work should investigate how to improve the adaptation for Braille tokenization, reasoning activation, and dealing with truncated outputs.
Besides, specific evaluation in Braille scenarios should be specialized, e.g., error recovery, user-centered usability, and safety-aware mechanisms must be quantified.


\bibliographystyle{IEEEtran}
\bibliography{ref}

\clearpage

\end{document}